\documentclass{article}
\usepackage{graphicx}%
\usepackage{multirow}%
\usepackage{amsmath,amssymb,amsfonts}%
\usepackage{amsthm}%
\usepackage{mathrsfs}%
\usepackage{algorithmicx}%
\usepackage{algpseudocode}%
\usepackage{listings}%
\usepackage{textcomp,multirow}
\usepackage{bm}
\usepackage{caption}

\DeclareCaptionLabelSeparator{bar}{\,\textbar\,}
\usepackage{url}
\usepackage{makecell}

\usepackage[superscript]{cite}
\usepackage{xcolor}
\usepackage{booktabs}
\usepackage{xurl}
\begin{document}
\newcommand{\ie}{{\it i.e.}}
\newcommand{\eg}{{\it e.g.}}

% Show comparison
% \newcommand{\zz}[2]{\par\noindent\textcolor{red!60!black}{\textbf{ZZ:} #1}\textcolor{blue}{\textbf{Original:} #2}}

% Show modified
\newcommand{\zz}[2]{#1}

% % Show original
% \newcommand{\zz}[2]{#2}

\title{Towards Lawful Autonomous Driving:\\
Deriving Scenario-Aware Driving Requirements from Traffic Laws and Regulations}

\author{
Bowen Jian$^{1,2,\dagger}$,
Rongjie Yu$^{1,2,*,\dagger}$,
Hong Wang$^{3}$,
Liqiang Wang$^{4}$,
Zihang Zou$^{5,*}$
\\[1ex]
\small
$^{1}$College of Transportation, Tongji University, Shanghai, 201804, China\\
\small
$^{2}$The Key Laboratory of Road and Traffic Engineering, \\
\small
Ministry of Education, Shanghai, 201804, China \\
\small
$^{3}$School of Vehicle and Mobility, Tsinghua University, Beijing, 100084, China\\
\small
$^{4}$College of Computer Science, University of Central Florida, Orlando, 32828, United States\\
\small
$^{5}$Optixway AI, Orlando, FL 32816, USA\\[0.5ex]
\small $^{\dagger}$These authors contributed equally: Bowen Jian, Rongjie Yu.\\
\small $^{*}$Correspondence: yurongjie@tongji.edu.cn; zzou@optixway.com\\
}

\maketitle
% abstract 
\begin{abstract}
{Driving in compliance with traffic laws and regulations is a basic requirement for human drivers, yet autonomous vehicles (AVs) can violate these requirements in diverse real-world scenarios. To encode law compliance into AV systems, conventional approaches use formal logic languages to explicitly specify behavioral constraints, but this process is labor-intensive, hard to scale, and costly to maintain. With recent advances in artificial intelligence, it is promising to leverage large language models (LLMs) to derive legal requirements from traffic laws and regulations. However, without explicitly grounding and reasoning in structured traffic scenarios, LLMs often retrieve irrelevant provisions or miss applicable ones, yielding imprecise requirements. To address this, we propose a novel pipeline that grounds LLM reasoning in a traffic scenario taxonomy through node-wise anchors that encode hierarchical semantics. On Chinese traffic laws and OnSite dataset (5,897 scenarios), our method improves law-scenario matching by 29.1\% and increases the accuracy of derived mandatory and prohibitive requirements by 36.9\% and 38.2\%, respectively.
We further demonstrate real-world applicability by constructing a law-compliance layer for AV navigation and developing an onboard, real-time compliance monitor for in-field testing, providing a solid foundation for future AV development, deployment, and regulatory oversight.
}
\end{abstract}

\section{Introduction}\label{intr}
Traffic laws and regulations are designed to enforce driving requirements for human drivers, refined over decades to ensure traffic safety and efficiency~\cite{geisslinger2023ethical, shalev2017formal,pek2020using}. Yet in recent years, as autonomous vehicles (AVs) move from closed-field testing to public roads, they have repeatedly struggled to comply with these long-established legal requirements, leading to traffic violations, license suspensions, and even fatal crashes.

In the past year alone, AV deployments have been involved in a growing number of law-compliance incidents:
Tesla’s China FSD trial was linked to repeated traffic violations, including lane-use violations and signal-control violations, exposing drivers to demerit-point penalties and possible license suspension~\cite{automotiveworld2025,carscoops2025,yahooautos2025,arenaev2025,globalchinaev2025}.
A Waymo driverless taxi was reported to make an illegal stop, blocking the road and hindering emergency crews responding to public crises~\cite{waymo_austin_block_emergency}.
Baidu’s Apollo Go robotaxi stopped within an intersection that caused long queues~\cite{sixthtone_apollogo_jams_2024,chinadailyhk_robotaxi_videos_2024}.
More severely, a fatality crash occurred in a barrier-marked work-zone diversion, where the Xiaomi vehicle failed to comply with work-zone traffic regulation and resulted in fatalities~\cite{reuters_xiaomi_su7_2025,yicaiglobal_xiaomi_su7_workzone_2025}.

These examples are not isolated corner cases but consequences of large-scale deployment and operational expansion in the absence of adequate regulatory oversight. On one hand, driver-assistance functions have reached mass adoption: more than 50\% of new cars sold are equipped with Level-2 systems~\cite{yicaiglobal_l2_2025}. Robotaxi services are also scaling rapidly. In the U.S., Waymo reported over 14 million robotaxi trips in 2025 alone, exceeding 1 million fully autonomous rides per month~\cite{waymo_year_in_review_2025,verge_waymo_14m_2025}. In China, Baidu’s Apollo Go surpassed 20 million cumulative rides by February 2026~\cite{chinadaily_apollogo_q4_2025}.
On the other hand, regulatory oversight is still catching up, often with delays and in response to incidents, through measures such as mandatory incident reporting, investigations, software recalls, and newly introduced regulatory pilot tests for L3 autonomy~\cite{nhtsa_sgo_crash_reporting,nhtsa_pe25013_2025,nhtsa_recall_25e084_2025,reuters_china_bans_terms_2025,reuters_china_ota_approvals_2025,chinadaily_pilots_L3_on_roads,ChinaMIIT2025}.

The primary reason AVs often fail to comply with traffic laws is that most models are trained in a data-driven fashion, learning driving behavior implicitly from data~\cite{hu2023planning,jiang2023vad, li2025end}. Unlike human drivers who must pass a knowledge exam and a road test to obtain a license, AV models have no analogous mechanism to explicitly encode traffic laws during training. Moreover, traffic-law compliance is inherently difficult because legal obligations are scenario-dependent and jurisdiction-specific. The applicability of a provision can vary substantially across countries, such as under right-hand versus left-hand traffic, or even across cities within the same country because of different posted limits. Compliance may further require jointly applying multiple provisions; for example, the legality of a left turn may depend simultaneously on signals, signs or road markings, and right-of-way rules. In addition, laws and regulations evolve over time, and revisions take effect on specific dates, requiring AV systems to switch promptly from outdated requirements to updated ones. 

To address this challenge, existing approaches translate legal provisions into formal, machine-checkable driving requirements, namely behavioral constraints that can be embedded into AV systems to monitor compliance. These methods typically adopt temporal-logic specifications~\cite{RN459,RN181,RN451,RN452,RN182,RN453,RN380} to specify when a rule applies and what constraints the vehicle must satisfy. Although effective in well-scoped settings, such logic-based rule bases depend heavily on expert-defined specifications, making them labor-intensive to construct, costly to maintain as regulations change, and difficult to scale to complex real-world scenarios.

\begin{figure}[t]
\centering
\includegraphics[width=\linewidth]{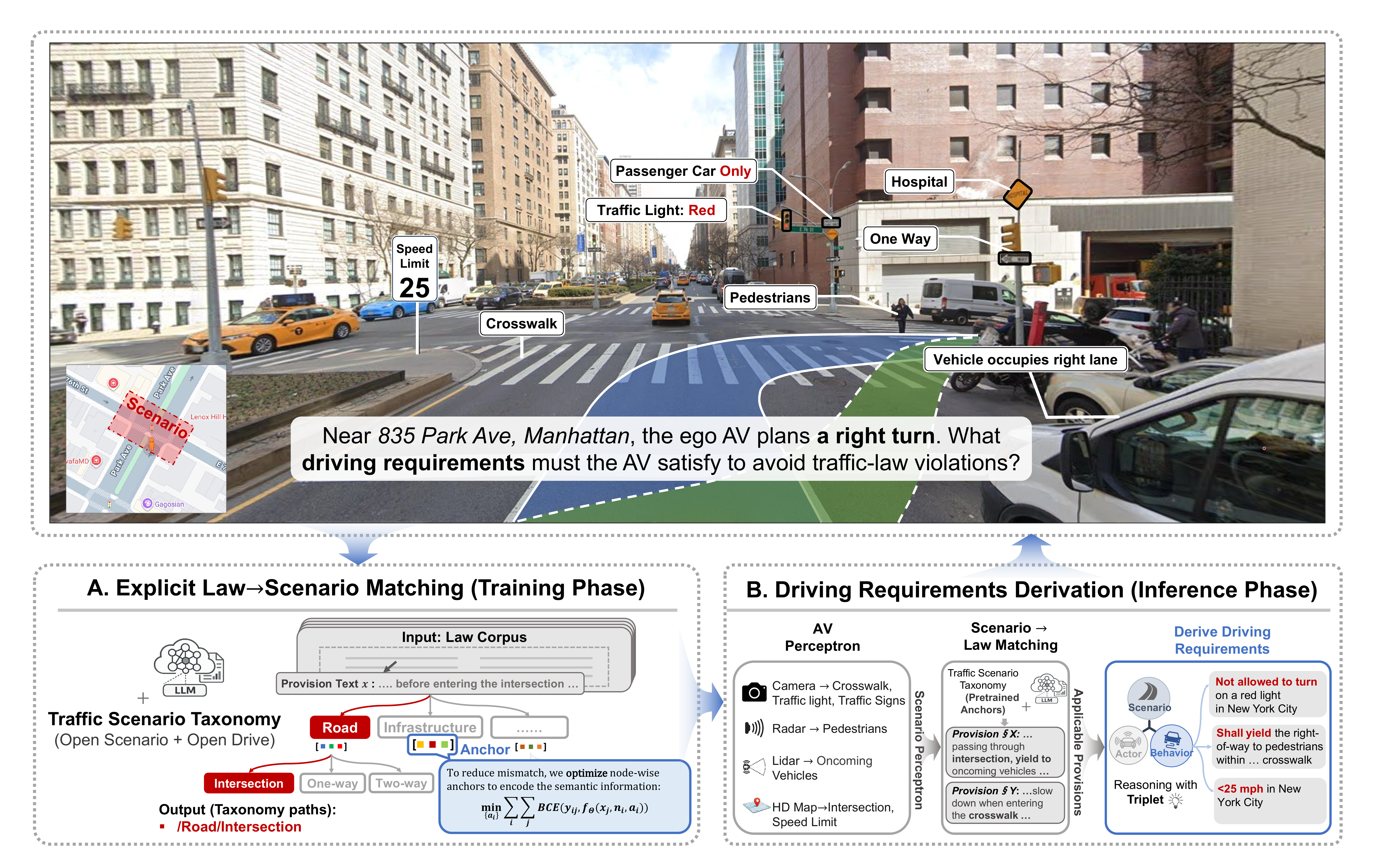}
\caption{\textbf{Overview of the pipeline to derive driving requirements.}
Our pipeline takes traffic scenarios as input and produces actionable driving requirements for AVs. During the training phase, a taxonomy with node-wise semantic indexes is learned to empower the LLM with better grounding in the associations between traffic scenarios and traffic laws. During inference, AV Perceptron is mapped to the learned taxonomy to retrieve applicable legal provisions. After that, a reasoning process derives actionable driving requirements from applicable provisions, supporting route choice, tactical planning, and operational control in AV systems. 
}
\label{fig_overview}
\end{figure}

With recent advances in artificial intelligence, especially large language models (LLMs)~\cite{brown2020language,achiam2023gpt,team2023gemini}, it is promising to develop an automated and scalable pipeline to derive legal driving requirements directly from traffic laws and regulations. Recent studies have explored using LLMs to incorporate traffic-legal text into AV decision-making, for example, via prompting~\cite{ouyang2022instructgpt, liu2023promptsurvey,zhou2025autovla}, fine-tuning~\cite{ding2023parameter,hu2022lora,lester2021power,fu2025orion}, or retrieval-augmented generation (RAG)~\cite{lewis2020rag,zhang2025safeauto,RN402,RN320}.
However, with diverse traffic scenarios and legal provisions, retrieving the applicable ones can easily trigger LLM hallucination~\cite{RN460,marcus2020next,ji2023survey,chen2023hallucination}.
This issue arises because the ``scenario'' is often provided as free-form text or raw perception labels, giving the model no canonical, structured handle to locate the case within the space of traffic scenarios. Without in-depth grounding, retrieval often follows surface keywords; for example, a ``work zone'' scenario may pull generic speed-limit provisions instead of the specific work-zone traffic-control and reduced-speed obligations, leading to mismatched requirements.

In this work, we propose a novel pipeline to derive driving requirements, namely behavioral constraints, from traffic legal provisions and regulations. 
Because driving requirements are not universally applicable but depend on scenario conditions, deriving them from legal text first requires identifying the conditions under which a provision applies. As shown in Fig.~\ref{fig_overview}, our pipeline consists of two components.

(1) To reduce ambiguity and hallucination, we implement explicit law--scenario matching by grounding both laws and scenarios in a traffic scenario taxonomy. Taxonomies are a well-established mechanism for semantic indexing that facilitates grounding and retrieval. Building on OpenDRIVE~\cite{ASAMOpenDRIVE} and OpenSCENARIO~\cite{ASAMOpenSCENARIO}, we unify their representations into a traffic scenario taxonomy and integrate it into the LLM through taxonomy-guided anchoring (TGA). 
During training, we attach a learnable soft prompt to each taxonomy node as its anchor, such that the anchor preserves the semantic information of that node. These anchors are then used to determine whether a legal provision contains the corresponding taxonomy concepts. During inference, the learned taxonomy anchors enable more accurate scenario grounding, which in turn improves the retrieval of the applicable provisions.

(2) Conditioned on the grounded scenario, our pipeline translates applicable legal provisions into actionable driving requirements through a structured Chain-of-Thought reasoning process.
These derived constraints are integrated into the AV system as multi-scale guidance, spanning from route choice to trajectory planning and vehicle control actuation. Specifically, the framework enforces strategic routing rules (\eg, turning prohibitions), tactical planning boundaries (\eg, yielding maneuvers), and operational control limits (\eg, speed and following distance thresholds). 
This hierarchical alignment operationalizes normative statutes as precise constraints for real-world driving.

We evaluate our method on Chinese traffic laws using the OnSite dataset~\cite{onsite_dataset}, a comprehensive benchmark comprising 5{,}897 scenarios. 
OnSite is widely recognized as a representative evaluation platform for high-level autonomous driving in China and has served as a key testing environment for the National Natural Science Foundation of China (NSFC) project validation. Our results show that TGA reduces LLM hallucinations relative to conventional LLM baselines and improves law--scenario understanding.

Beyond the Chinese benchmark, we further demonstrate the practical applicability of the proposed pipeline in two additional settings. First, we construct a law-compliance layer based on structured map data across nine major robotaxi operating regions in the United States and China. Second, we deploy the pipeline in field testing as an onboard, real-time compliance monitor for AV operation. Together, these results demonstrate the scalability of the proposed pipeline for deriving scenario-based legal requirements to support lawful AV development, deployment, and regulatory oversight.

\section{Results}
In the following, we first explore whether the LLM can be accurately grounded in the inference about the applicable scenarios of traffic laws, by performing a law--scenario matching task (Section~\ref{law_scenario_matching}). We subsequently evaluate the consistency between the driving requirements derived by LLMs and the legal grounding (Section~\ref{derivation}). Additionally, we promote the pipeline to two main applications, including augmenting digital navigation with legal semantics which supports AVs to plan compliance behaviors (Section~\ref{res_layer}), and developing an offline compliance monitor, demonstrating regulatory oversight for AVs in the real world (Section~\ref{res_field}).

\subsection{Explicit law--scenario matching}
\label{law_scenario_matching}
\noindent\textbf{Scenario taxonomy.} We integrate three international standards -- OpenDRIVE~\cite{ASAMOpenDRIVE}, OpenSCENARIO~\cite{ASAMOpenSCENARIO}, and ISO 34504~\cite{ISO34504} -- into a unified scenario taxonomy that organizes traffic scenarios in a hierarchical taxonomy, spanning road, infrastructure, traffic management, environment, objects, and digital information. 
The {\it Road} captures attributes such as geometry, cross-sectional profiles, and longitudinal alignment features. {\it Infrastructure} includes elements such as road markings, traffic signs, traffic lights, and central dividers. {\it Traffic management} covers temporary or administrative factors, including temporary traffic control and traffic events. {\it Environment} describes external conditions such as weather, visibility, and time of day. The {\it Objects} represent traffic participants and their motions. {\it Digital information} refers to messages shared via the Internet. Overall, the taxonomy contains 227 nodes, each assigned a scenario tag given by its root-to-node path (\eg, \texttt{/Road/Intersection/T-intersection}).

\noindent\textbf{Traffic laws.}
We construct a traffic-law corpus with China as the primary jurisdiction. The corpus includes five authoritative documents: four national traffic laws and one national standard. These documents are typically organized into paragraphs for enforcement convenience, and a single paragraph may enumerate similar driving requirements that apply across multiple traffic scenarios. Following this normative structure, we first segment each document by paragraph. 
If a paragraph covers multiple scenarios, we further split it so each provision describes one scenario defined by a distinct combination of taxonomy nodes.
After refinement, the corpus contains 310 legal provisions.

\noindent\textbf{Dataset.}
We annotate each legal provision by matching it to the proposed scenario taxonomy and assigning the set of applicable scenario tags that characterize its driving context. These annotations serve as ground truth for the law–scenario matching task. 

\noindent\textbf{Evaluation.}
Given the limited number of provisions, we adopt five-fold cross-validation. In each fold, we split the annotated provisions into training and validation sets and report results averaged across folds.
Because each legal provision may correspond to multiple scenario tags, law–scenario matching is a multi-label classification task. We report precision, recall, and F1-score. Here, low precision indicates mismatches involving non-applicable tags, whereas low recall reflects mismatches involving missing applicable tags. F1-score provides a balanced view of overall matching performance.

\noindent\textbf{Baseline.}
We selectively compare three representative LLM settings for law--scenario matching tasks. (1) {\it Inference-only}: a pre-trained LLM relies solely on its internal knowledge to infer the applicable scenario tags for a provision. (2) {\it Fine-tuning}: the model is trained on annotated provisions so that law–scenario matching knowledge is encoded into network parameters. (3) {\it RAG}: the model retrieves external knowledge from an official scenario categorization document~\cite{ISO34504} at inference time to support tag prediction. More details can be found in Appendix~A.

\noindent\textbf{Configuration.}
All experiments use the same foundation model, Qwen3-8B~\cite{yang2025qwen3}. 
For our approach, we introduce learnable taxonomy anchors that encode node-specific knowledge to improve law--scenario matching. Each anchor is implemented as a node-wise soft prompt of length 20 tokens and is trained using Adam~\cite{kingma2015adam} with a learning rate of 0.003.

\noindent\textbf{Performance Comparison.} As shown in Table~\ref{tab_res1}, the proposed method achieves the highest F1-score, outperforming the strongest baseline by 29.1\%, whereas other methods exhibit a clear trade-off between precision and recall. TGA improves \emph{both} sides of the mismatch spectrum: taxonomy anchors inject node-specific semantics that sharpen the decision boundary between similar tags, reducing mismatches from predicting non-applicable tags (higher precision), while the hierarchical organization encourages consistent coverage of related scenario factors, reducing mismatches from missing applicable tags (higher recall).

\begin{table}[t]
\centering
\caption{\textbf{Performance comparison for law--scenario matching}. Low precisions indicate over-matching irrelevant tags. Low Recalls imply LLM omits tags. }
\label{tab_res1}
{\fontsize{8pt}{9pt}\selectfont{}
    \begin{tabular}{p{3cm}p{4cm}ccc}
    \toprule
    \textbf{Category} & \textbf{Method} & \textbf{Recall} & \textbf{Precision} & \textbf{F1} \\
    \midrule
    \multirow{3}{*}{Inference-only}
      & Zero-shot & \textbf{0.956} & 0.193 & 0.304 \\
      & Zero-shot+thinking & 0.939 & 0.230 & 0.348 \\
      & Zero-shot+knowledge & 0.947 & 0.192 & 0.305 \\
    \cmidrule{1-5}
    Fine-tuning & Lora & 0.390 & \textbf{0.873} & 0.441 \\
    \cmidrule{1-5}
    \multirow{2}{*}{RAG}
      & VanillaRAG & 0.925 & 0.397 & \underline{0.513} \\
      & GraphRAG & 0.866 & 0.037 & 0.070 \\
    \cmidrule{1-5}
      & TGA (Ours)  & \underline{0.953} & \underline{0.763} & \textbf{0.804} \\
    \bottomrule
    \end{tabular}
}
\end{table}

We further analyze the inference-only setting to explain why zero-shot prompting attains high recall yet low precision. The zero-shot baseline frequently over-matches non-applicable tags, likely because the pre-trained model relies on real-world co-occurrence rather than the applicability conditions stated in legal documents. For example, provisions that merely mention ``traffic lights'' are often matched to ``intersection'' or ``crosswalk'' even when those scenarios are not specified. Enabling thinking mode slightly improves precision by encouraging definition-based reasoning and verification against the provision text. In contrast, adding textual knowledge yields limited gains, likely because many coarse correlations are already present from pre-training.

The fine-tuned model achieves higher precision by better filtering non-applicable tags, but its recall drops substantially, indicating that substantial applicable tags are missed. This performance is consistent with LoRA-style adaptation, which shifts the model toward the fine-tuning distribution. Since the dataset is highly imbalanced and each provision corresponds to only a small subset of tags, the fine-tuned model becomes overly conservative, favoring negative predictions to avoid false positives and thereby omitting valid tags.

RAG performance depends on whether retrieval provides fine-grained, task-aligned evidence for deciding applicability. Vanilla RAG uses embedding-based semantic similarity to retrieve targeted passages relevant to the given provision–tag query, yielding the second-best performance. In contrast, GraphRAG aggregates broader context via community detection and graph traversal~\cite{RN320}, which often introduces redundant or weakly relevant content and degrades performance below the zero-shot baseline. Empirically, GraphRAG retrieves about 1{,}300 tokens per query, compared to roughly 300 tokens for Vanilla RAG, and the additional context frequently includes section-level introductions and irrelevant definitions that dilute task-relevant signals.

The results above demonstrate the effectiveness of the proposed method. We further conduct an ablation study in Table~\ref{tab:ablation} to analyze how different anchor designs influence performance on the law--scenario matching task.

\begin{table}[h]
\caption{\textbf{Ablation study on law--scenario matching}}
\label{tab:ablation}
{\fontsize{8pt}{9pt}\selectfont
\begin{tabular}{p{7.4cm}ccc}
\toprule
\textbf{Setting} & \textbf{Recall} & \textbf{Precision} & \textbf{F1} \\
\midrule

\multicolumn{4}{l}{\textbf{A. Explicit vs.\ Implicit Knowledge}} \\
\midrule
Handcrafted prompts & 0.918 & 0.715 & 0.764 \\
Learnable Anchor & \textbf{0.953} & \textbf{0.763} & \textbf{0.804} \\
\midrule

\multicolumn{4}{l}{\textbf{B. Universal vs.\ Local Knowledge}} \\
\midrule
Universal anchor & 0.944 & 0.161 & 0.260 \\
Node-wise anchor & \textbf{0.953} & \textbf{0.763} & \textbf{0.804} \\
\midrule

\multicolumn{4}{l}{\textbf{C. Learning Budget}} \\
\midrule
5 tokens & 0.943 & 0.762 & 0.801 \\
20 tokens & \textbf{0.953} & 0.763 & \textbf{0.804} \\
50 tokens & \underline{0.952} & 0.756 & 0.801 \\
100 tokens & 0.933 & \textbf{0.773} & \underline{0.802} \\
200 tokens & 0.896 & \underline{0.770} & 0.790 \\
\bottomrule
\end{tabular}
}
\end{table}

\noindent\textbf{Explicit vs.\ implicit knowledge.}
To assess whether hierarchical scenario knowledge is better injected explicitly or learned implicitly, we compare handcrafted prompts with our learned soft prompts. The handcrafted prompts provide node-specific textual definitions obtained via few-shot reasoning over the law corpus for each taxonomy node. Despite requiring over 17$\times$ more input tokens (788 vs.\ 45), handcrafted prompts still underperform learned anchor methods, indicating that implicit knowledge learning is both more token-efficient and more effective at capturing the latent semantics needed for accurate matching.

\noindent\textbf{Universal vs.\ local knowledge.}
To evaluate whether node-wise anchors are necessary, we replace all local anchors with a single universal anchor shared across the taxonomy. This change leads to a substantial performance drop, indicating strong semantic heterogeneity across scenario tags, \ie, tags depend on different textual cues. As a result, a universal anchor cannot provide the tag-specific representations needed to separate closely related tags, weakening decision boundaries and degrading matching accuracy.

\noindent\textbf{Learning budget.}
To examine how learning capacity affects performance, we vary the anchor length from 5 to 200 tokens. Very short anchors ($<10$ tokens) are capacity-limited and create a semantic bottleneck, failing to capture enough knowledge needed for each tag. In contrast, overly long anchors degrade performance, likely due to harder optimization and increased noise under limited supervision. Mid-range lengths are more stable, with the best performance attained at 20 tokens, which provides sufficient capacity to encode tag-specific semantics without incurring optimization instability.

\subsection{Deriving driving requirements across scenarios}
\label{derivation}
\noindent\textbf{Datasets.}
We use OnSite~\cite{onsite_dataset}, a publicly available Chinese traffic scenario dataset, as a test set for driving-requirement derivation. OnSite aggregates scenarios from two sources: high-risk driving video segments collected through naturalistic driving studies, and roadside cameras deployed in China’s High-Level Automated Driving demonstration areas. In total, OnSite contains 5,897 scenarios represented in compliance with the OpenX standards~\cite{ASAMOpenDRIVE, ASAMOpenSCENARIO}.

To systematically derive driving requirements, we first decompose the 5,897 continuous scenarios into 104,608 discrete, segment-level ``regulatory analysis units'' based on local feature transitions (\eg, changes in road markings). Since identical combinations of scenario features should map to the same legal provisions and driving requirements, we further consolidate these units into 896 unique scenario-tag combinations. These 896 combinations then serve as the core analytical test samples for driving-requirement derivation.

Building upon the grounded law–scenario matching, we established a rigorous benchmark by manually deriving the applicable legal provisions into driving requirements. Specifically, to distinguish different constraints on AV's expected behaviors, the requirements are classified into two normative categories: prohibitive behaviors, which define the negative space of actions (\ie, what must be avoided), and mandatory behaviors, which prescribe positive obligations (\ie, what must be executed). Crucially, this benchmark extends beyond static, ego-centric restrictions (\eg, a strict parking prohibition and speed reduction mandate within a work zone) to encompass complex, interaction-conditioned dynamics. For example, a prohibition on overtaking may be dynamically triggered by the specific state of a preceding vehicle signaling a left turn. Capturing these multi-agent relational constraints ensures that our evaluation rigorously reflects the highly interactive nature of real-world compliance.

\noindent\textbf{Evaluation Metrics.}
Three complementary text-based metrics are adopted to produce a comprehensive measurement for derived textual driving requirements. The 1-gram F1~\cite{RN221} measures lexical correctness and reflects whether the types of behaviors are identified. The 2-gram F1~\cite{RN221} captures the sequence and dependency relations among multiple behaviors through short-range word-order. And BERTScore F1~\cite{zhang2020bertscore} rewards paraphrases that preserve legal meaning even when surface wording differs, by calculating semantic similarity using contextualized embeddings. 

\noindent\textbf{Baselines.}
Following the experimental setup of the law–scenario matching task, we compare three representative LLM settings for driving-requirement derivation based on a unified foundation model (Qwen3-8B). (1) {\it Inference-only}: a pre-trained LLM relies solely on its intrinsic knowledge to infer normative requirements. (2) {\it Fine-tuning}: the model is trained on the traffic-law corpus so that regulatory knowledge is encoded into network parameters via a next-token-prediction paradigm. (3) {\it RAG}: the model automatically retrieves relevant legal provisions at inference time to conditionally guide its subsequent behavioral reasoning. 

\noindent\textbf{Configuration.}
To operationalize this derivation process, we constructed a structured legal knowledge base where each provision is tightly bound to its explicit applicability scope, inheriting the mapping foundations established in the law–scenario matching task. Leveraging the underlying scenario taxonomy, we implemented a hierarchy-aware retrieval strategy. This multi-level formulation ensures comprehensive coverage by enabling the simultaneous extraction of both overarching statutory principles (general provisions) and granular, context-dependent rules (specific provisions). Subsequently, we employed a Chain-of-Thought (CoT) reasoning framework to systematically translate these retrieved legal texts into driving requirements. 

\begin{figure}[h]
\centering
\includegraphics[width=\linewidth]{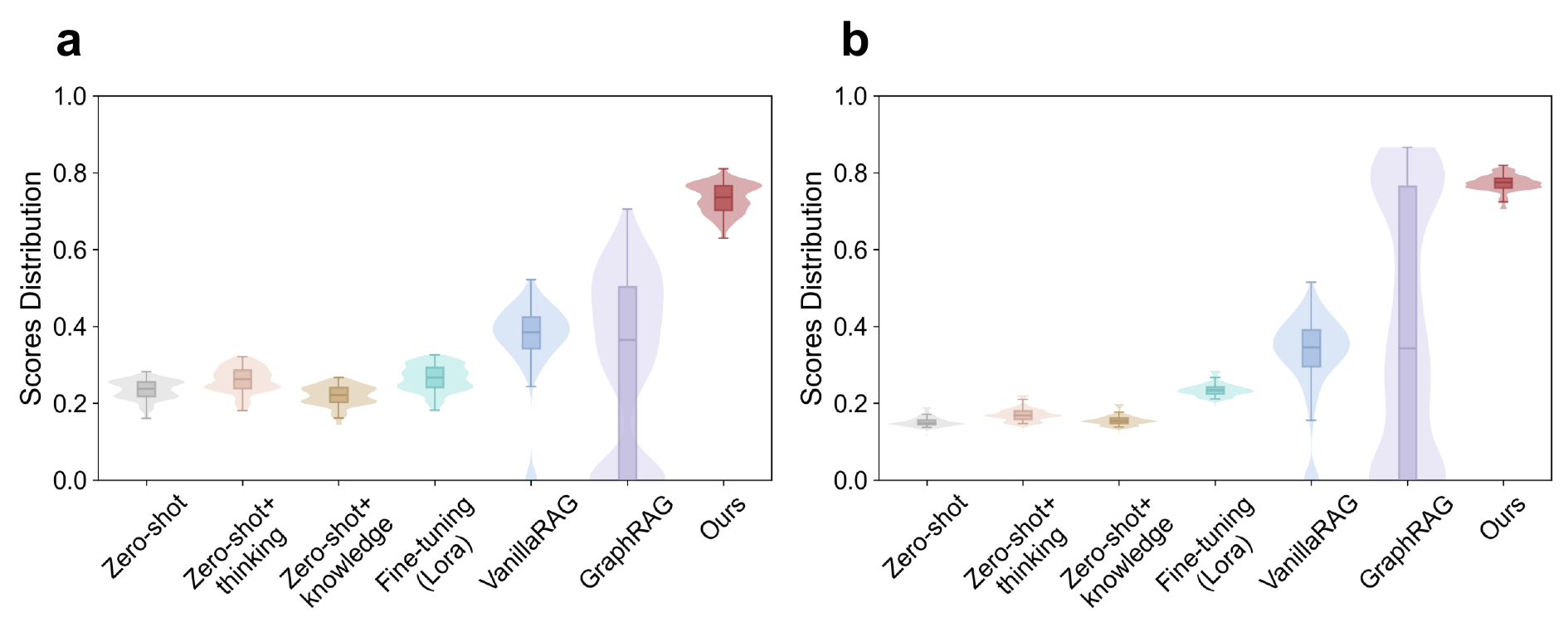}
\caption{\textbf{Performance distribution across scenarios for driving-requirements derivation.} 
The figures \textbf{a} and \textbf{b} report the 1-gram F1 score distributions for derived mandatory and prohibitive behaviors, separately. The scores are evaluated over 896 unique scenario-tag combinations in the OnSite dataset.
For each method, scores across all combinations are aggregated and visualized using violin plots, where the violin width indicates the relative frequency of specific score values and the vertical extent reflects the overall range of performance across scenarios. A larger vertical spread suggests higher sensitivity to scenario variations. Compared with baseline methods, our approach achieves consistently higher mean performance with a markedly more concentrated score distribution, indicating superior accuracy as well as stability across diverse scenarios.
}
\label{fig_violin}
\end{figure}

\noindent\textbf{Performance Comparison.}
Since mandatory and prohibitive behaviors impose distinct types of constraints on AVs, we quantitatively evaluate the derivation accuracy in Table \ref{tab_res_derivation}, separately, with the performance variance across diverse scenarios visualized via violin plots (Fig.~\ref{fig_violin}). Our taxonomy-grounded pipeline demonstrates decisive superiority over conventional baselines, achieving substantial relative gains of 36.9\% and 38.2\% in type accuracy (\ie, 1-gram F1) for mandatory and prohibitive constraints, respectively, while maintaining robust performance across diverse scenarios.  This dual advantage stems from utilizing the scenario taxonomy as a semantic anchor to precisely link physical driving environments with applicable legal provisions. In contrast, limitations in capturing these associations have been exposed in conventional methods:

\begin{table}[h]
\centering
\caption{\textbf{Performance of deriving driving requirements from tagged scenarios.} High 1-gram F1 implies correct behavior types. High 2-gram F1 indicates precise capture of the dependency relations among multiple behaviors. High BERTScore F1 means derived behaviors and ground truth are semantically similar.}
\label{tab_res_derivation}
{\fontsize{8pt}{9pt}\selectfont
    \begin{tabular}{llccc}
    \toprule
    \textbf{Category} & \textbf{Method} & \textbf{1-gram F1} & \textbf{2-gram F1} & \textbf{BERTScore F1} \\
    \midrule

    \multicolumn{5}{l}{\textbf{A. Mandatory behaviors}} \\
    \midrule
    \multirow{3}{*}{Inference-only}
     & Zero-shot & 0.234 & 0.045 & 0.310 \\
     & Zero-shot+thinking & 0.261 & 0.054 & 0.339 \\
     & Zero-shot+knowledge & 0.219 & 0.047 & 0.329 \\
    \midrule
    Fine-tuning & Lora & 0.265 & 0.075 & \underline{0.356} \\
    \midrule
    \multirow{2}{*}{RAG}
     & VanillaRAG & \underline{0.364} & 0.133 & 0.317 \\
     & GraphRAG & 0.314 & \underline{0.176} & 0.116 \\
    \midrule
     & TGA (Ours) & \textbf{0.733} & \textbf{0.592} & \textbf{0.590} \\
    \midrule

    \multicolumn{5}{l}{\textbf{B. Prohibitive behaviors}} \\
    \midrule
    \multirow{3}{*}{Inference-only}
     & Zero-shot & 0.152 & 0.037 & 0.360 \\
     & Zero-shot+thinking & 0.171 & 0.032 & 0.316 \\
     & Zero-shot+knowledge & 0.156 & 0.035 & 0.340 \\
    \midrule
    Fine-tuning & Lora & 0.235 & 0.058 & \underline{0.370} \\
    \midrule
    \multirow{2}{*}{RAG}
     & VanillaRAG & 0.321 & 0.127 & 0.284 \\
     & GraphRAG & \underline{0.390} & \underline{0.279} & 0.180 \\
    \midrule
     & TGA (Ours)  & \textbf{0.772} & \textbf{0.675} & \textbf{0.695} \\
    \bottomrule
    \end{tabular}
}
\end{table}

The zero-shot baseline exhibited notably low accuracy, indicating that the pre-trained LLM has not inherently internalized the driving requirements of traffic regulations. To further investigate this deficiency, we conducted a supplementary experiment using standard driver's license examination Q\&As (detailed in Appendix~C). The results indicate that the pre-trained LLM exhibits strong memorization of lawful behaviors when presented with textbook-style questions highly consistent with raw traffic-law contents. However, pre-training alone is insufficient for grounding normative requirements in complex, real-world driving environments.

Using a thinking mode slightly improved performance, as the model attempted to concentrate on partial tokens to infer driving requirements. For instance, when presented with a three-lane freeway and a single solid white line, the model primarily focused on the token ``freeway'' to infer relevant behaviors. However, such simple reasoning remains tag-isolated and fails to capture the combinatorial dependencies between multiple scenario features, limiting the ability to map complex scenario-law-behavior relationships. Incorporating definitions of driving requirements without the linkage between scenarios and laws provided minimal additional benefit.  

Fine-tuning yields marginal performance gains by better adhering to the syntactic structure of raw legal texts. However, statically embedding regulatory knowledge fails to equip the model for the dynamic variability of real-world scenarios. Without explicit scenario-law anchored reasoning, the model struggles to dynamically infer precise driving requirements, resulting in performance volatility.

RAG paradigms offered moderate gains by explicitly conditioning inference on retrieved provisions, yet inherent architectural limitations persist. Vanilla RAG is fundamentally bottlenecked by the profound semantic gap between concrete physical scenario descriptions and abstract legal texts, rendering standard dense vector retrieval ineffective at capturing their misaligned lexical correspondences. Conversely, while Graph RAG attempts to capture structural legal associations, it relies on undirected, disconnected topologies that fundamentally clash with the hierarchical nature of the traffic scenario taxonomy~\cite{zhuang2025linearrag}. This topological misalignment introduces severe informational noise, causing high performance volatility (see violin plots in Fig.~\ref{fig_violin}).

\subsection{Enhancing AV navigation with a law-compliance layer}
\label{res_layer}
Digital navigation maps are foundational to the strategic routing and tactical operation of AVs. 
However, existing semantic maps primarily encode physical geometry, topological connectivity, and basic traffic-control information, but do not capture the nuanced, scenario-dependent driving requirements mandated by traffic laws. 
To bridge this gap, we introduce a pipeline that augments semantic map representations with a structured law-compliance layer, transforming conventional road networks into law-aware navigation environments.

\begin{figure}[h]
\centering
\includegraphics[width=\linewidth]{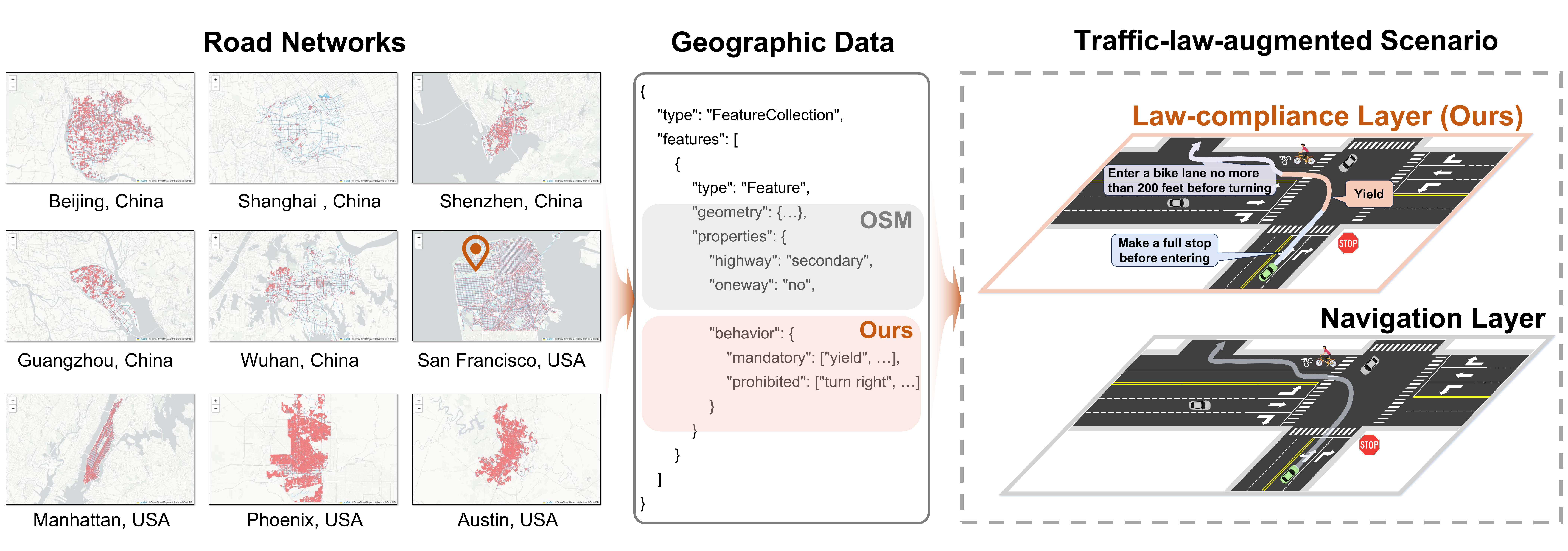}
\caption{\textbf{Attaching law-compliance layer into OpenStreetMap}. 
We built law-compliance layers based on OpenStreetMap road network data across multiple regions. 
Conventional traffic scenario representations primarily model static elements (\eg, road geometry and markings) and dynamic elements (\eg, traffic participants and their motions), following structured formats like GeoJSON. In addition, our pipeline augments these representations by introducing a dedicated law-compliance layer that encodes traffic-law-derived behavioral constraints governing AV actions. This law-compliance layer enables scenarios to explicitly capture driving requirements, supporting not only lawful driving for en-route AVs, but also the traffic enforcement criteria for regulatory oversight.
}\label{fig_map}
\end{figure}

We instantiate this pipeline on OpenStreetMap (OSM) road networks. 
First, we derive a semantic navigation map from the OSM road graph. We then add a law-compliance layer by attaching driving requirements derived by our TGA method to the corresponding map entities. 
As shown in Fig.~\ref{fig_map}, these requirements are encoded as GeoJSON attributes that specify mandatory and prohibitive behaviors, enabling law-aware navigation and downstream planning.

To demonstrate real-world applicability, we constructed over 1{,}500~km of law-augmented roadway networks spanning nine major robotaxi operation regions across the United States and China. To validate that the law-compliance layer can be exercised during navigation, we instantiate lane-based driving scenarios on these maps and simulate background traffic in SUMO~\cite{sumo_official_cite}, allowing us to evaluate law-compliant behavior under realistic interactions.
While we use open-source OSM as a proof of concept, the workflow is modular and can be adapted to other lane-level map providers (\eg, Google Maps or Gaode Maps) via customized feature extraction, enabling the same law-compliance layer to be deployed on standard navigation maps. Details can be found in Appendix C.

\subsection{In-field empirical compliance monitoring of AVs}
\label{res_field}
\begin{figure}[!h]
\centering
\includegraphics[width=\linewidth]{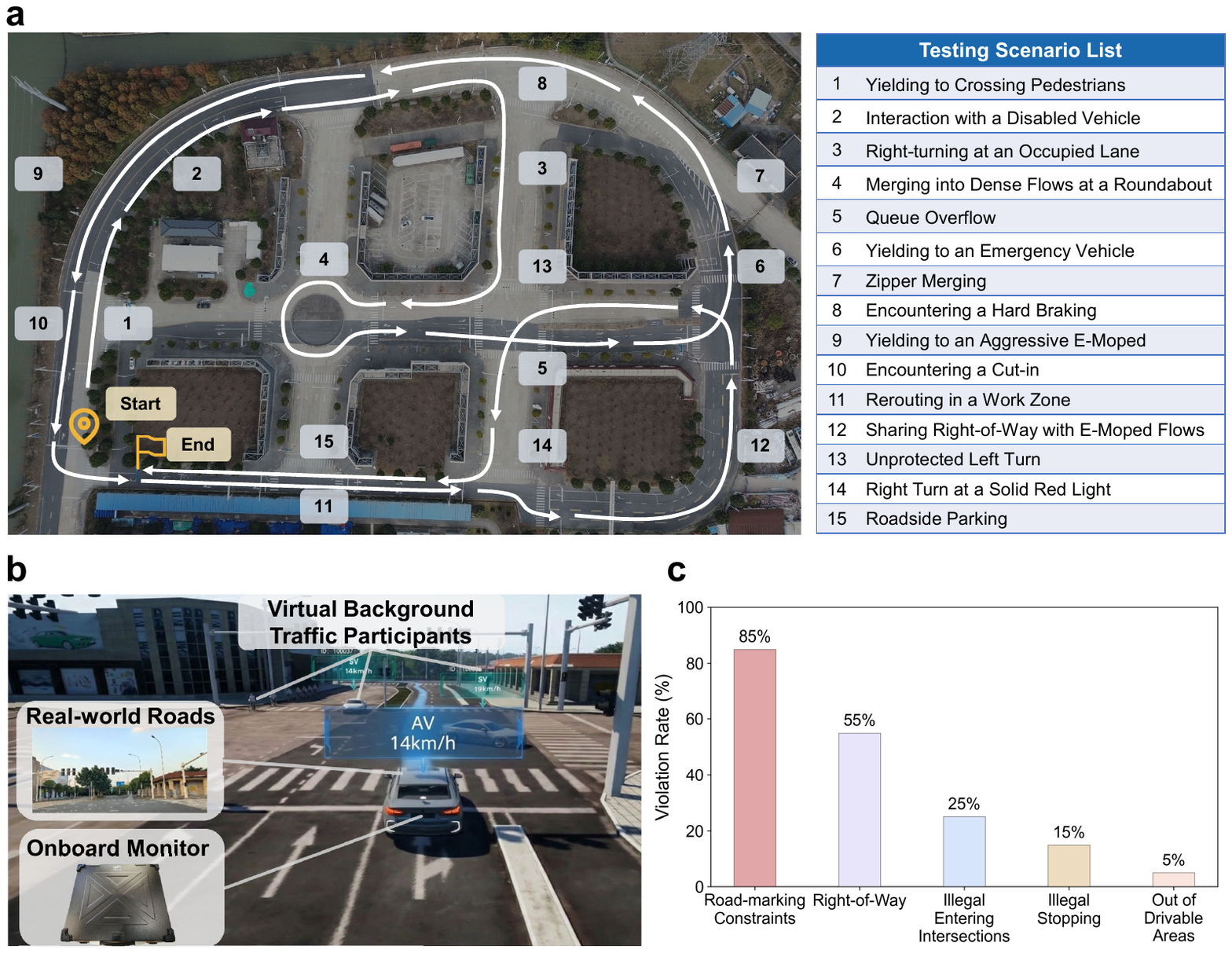}
\caption{\textbf{Empirical compliance monitoring for AVs in field testing}. 
\textbf{a.} Test scenarios layout for real-world evaluation. The 15 scenarios are designed by carefully combining road markings with trigger-based background vehicles to induce AV violations, referencing real-world urban road cases characterized by strong interactions and high-probability violations committed by human drivers.
\textbf{b.} A demonstration of the compliance monitor in a mixed-reality setting. The AV interacts with virtually injected background traffic participants superimposed onto real-world roads. Our monitor is equipped in the AV's trunk, which receives real-time trajectories from both the ego and background participants and then identifies traffic violations.
\textbf{c.} Law-compliance monitoring results of 20 AV systems. The violation rate, \ie, the vertical axis, denotes the percentage of AVs that committed a specific violation, with the horizontal axis representing different violation types. All monitoring results in the field test have been verified and proven to be accurate by experts. The results indicate our monitoring system can capture violations triggered by both ego-vehicle behaviors and complex multi-agent interactions.
}\label{fig_field_test}
\end{figure}

To extend evaluation beyond offline verification, we implemented our pipeline as an onboard, real-time compliance monitor. In a mixed-reality setting, background traffic is virtually injected onto real roads, and the monitor checks observed trajectories against scenario-specific legal primitives derived from traffic laws to flag violations with traceable accountability. As shown in Fig.~\ref{fig_field_test}, we validated the system in a large-scale national autonomous driving competition with 20 AV systems from leading research institutions, evaluated on high-complexity urban scenarios in China.

The monitoring reveals two dominant failure modes: misunderstanding right-of-way semantics and violating road-marking constraints. More than half of the tested systems failed to yield appropriately to priority entities, including emergency vehicles, oncoming through-traffic at unsignalized intersections, and pedestrians at marked crosswalks. These behaviors suggest that many AV stacks encode traffic participants primarily as physical objects (\eg, bounding boxes) without explicitly representing legally defined priority relations. In addition, 85\% of the systems performed illegal maneuvers across lane markings, including crossing double solid yellow lines, changing lanes across solid white lines, and sustained straddling of dashed lane boundaries. Together, these findings motivate law-aware occupancy and planning frameworks that distinguish physical feasibility from legal permissibility.
Two replays of these violations can be found in Supplementary Movies 1 and 2.

\section{Discussion}
\label{disc}
In this work, we developed a novel pipeline for deriving scenario-specific driving requirements from traffic laws and regulations. Conventional formal-logic approaches are difficult to scale, while LLMs often suffer from imprecise retrieval and missing applicable provisions. By grounding LLM reasoning through taxonomy-guided anchoring, our framework improves both law--scenario matching and the accuracy of derived driving requirements. We further validated the proposed architecture by constructing a law-compliance layer based on structured map data and by deploying an onboard real-time compliance monitor for autonomous driving systems, supporting a scalable path toward lawful autonomous driving and regulatory oversight.

To truly facilitate lawful autonomous driving, the framework established in this work should be further advanced at three levels: the model level, the decision-making level, and the regulatory level. At the model level, explicit compliance requirements should be more tightly integrated into data-driven end-to-end (E2E) AV architectures, potentially through advances in post-training methods or cross-modal fusion with semantic models such as vision--language models (VLMs). At the decision-making level, future systems should support more flexible mechanisms for applying behavioral constraints, especially in safety-critical situations where temporary rule deviations may need to be justified. At the regulatory level, the current traffic regulatory system should continue evolving beyond its human-centered design as AV deployment scales and transportation systems become increasingly automated.

The implications of this work may also extend beyond autonomous driving. As AI systems are increasingly deployed in domains governed by legal, ethical, and regulatory rules, aligning model behavior with normative requirements is becoming an important challenge. In this sense, the problem studied here reflects a broader question in AI safety and alignment: how human-written rules can be translated into machine-interpretable constraints that guide system behavior in context. By providing a framework for deriving such constraints from legal text, this work offers one possible direction toward more norm-aware AI systems in real-world applications.

\section{Methods}
\label{sec:method}

\subsection{Scenario Taxonomy Representation} 
The traffic scenario taxonomy is represented as a collection of root-to-node paths (\eg, \texttt{/Road/Intersection/T-intersection}), where each node on a path corresponds to a unique scenario tag used for taxonomy indexing. This representation captures the taxonomy hierarchy through the parent--child relations along each path. In addition, the taxonomy specifies that some sibling tags are mutually exclusive. For example, \texttt{single-lane} and \texttt{two-lanes} are mutually exclusive sibling tags. By contrast, \texttt{stop line} and \texttt{crosswalk} are not exclusive and can co-occur.

\subsection{Taxonomy-Guided Anchoring} 
Given a legal provision $x_j$, the task is to determine which taxonomy tags $n_i$ are applicable to $x_j$.
% \paragraph{Semantic anchoring for taxonomy nodes.} 
For each pair $(x_j, n_i)$, a pre-trained LLM is conditioned on the raw text of $x_j$ and the tag of $n_i$ to produce a binary yes/no decision, as shown in Table~\ref{textual_prompt}. Each taxonomy node $n_i$ is associated with an \emph{anchor} $a_i\in\mathbb{R}^{L\times d}$, implemented as a learnable soft prompt, where $L$ is the prompt length and $d$ is the embedding dimension. For provision-to-tag matching, the frozen LLM is conditioned on both the textual prompt and the node-specific anchor. The anchors $\{a_i\}$ are optimized using binary cross entropy,

\begin{equation}
\min_{\{a_i\}} \sum_{i}\sum_{j}\mathrm{BCE}\big(y_{ij}, f_{\bm{\Theta}}(x_j, n_i, a_i)\big),
\end{equation}
where $y_{ij}\in\{0,1\}$ indicates whether legal provision $x_j$ matches taxonomy tag $n_i$, and $f_{\bm{\Theta}}(x_j, n_i, a_i)$ denotes the predicted probability of the \texttt{Yes} label.

\begin{table}[htbp]
\centering
\caption{\textbf{Textual prompt for law--scenario matching}}
\label{textual_prompt}
{\fontsize{8pt}{9pt}\selectfont{}
\begin{tabular}{p{\linewidth}}
\toprule
\textbf{System:} You are an expert in traffic laws. Judge whether the following statement is correct. Just answer ``Yes'' or ``No''. \\
\textbf{User:} The traffic law ``\{law text\}'' describes a scenario that includes ``\{tag\}''. \\
\bottomrule
\end{tabular}
}
\end{table}

\subsection{Driving-requirement derivation}

\paragraph{Traffic-law retrieval.}
We retrieve applicable legal provisions from a knowledge base according to their applicability conditions. Specifically, the raw text of each legal provision is treated as the value $V$, and its associated taxonomy tags are treated as the key set $K$. Given a traffic scenario, all active scenario tags are extracted and augmented with their ancestor tags in the taxonomy to form the query set $Q$, so that both general (\eg, \texttt{road}) and specific (\eg, \texttt{urban road}) scenario features are included. A provision $V$ is retrieved if its applicability conditions are fully satisfied by the given scenario, corresponding to the subset condition $K \subseteq Q$.

\paragraph{Driving-requirement derivation via CoT.}
Because the retrieved provisions $V$ often entangle driving requirements with scenario descriptions, we use the Qwen3-8B guided by a two-step CoT process to derive actionable requirements. First, the CoT prompt instructs the LLM to categorize each provision into mandatory or prohibitive requirements by identifying normative indicators (\eg, modal verbs such as ``should'' and prescriptive phrases like ``not allowed''). The corresponding behavioral descriptions are then derived and assigned to their categories. By processing the retrieved provisions sequentially, the derived mandatory and prohibitive behaviors are aggregated separately to form the final set of driving requirements.

\section{Data availability}
Raw documents of Chinese traffic laws and regulations are available at: \url{https://flk.npc.gov.cn/index}. Raw driver handbooks for different U.S. states are available at: \url{https://driving-tests.org/}. 
% We also provide the processed legal provisions at \url{https://github.com/BowenJIAN/TrafficRule_compliance}, along with the source code.
The OnSite dataset used to validate the proposed method is public at \url{https://onsite.com.cn/#/dist/benchmarkLeaderBoard}. 
The map data used in this paper is extracted from OpenStreetMap, available at \url{https://www.openstreetmap.org/}. 
In-field testing data are available from the corresponding author, Rongjie Yu, on reasonable request.

% \section{Code availability}
% Source code is available at \url{https://github.com/BowenJIAN/TrafficRule_compliance}

\clearpage
\bibliographystyle{unsrt}
\bibliography{NC_format/sn-bibliography}

\section{Acknowledgments}
This study was sponsored by the Chinese National Natural Science Foundation (NSFC 5253000527).

\end{document}